\documentclass[11pt]{article}
\usepackage{acl2016}
\usepackage{times}
\usepackage{latexsym}
\usepackage[fleqn]{amsmath}
\usepackage{amssymb}
\usepackage{gensymb}
\usepackage[pdftex]{graphicx}
\usepackage{color}
\DeclareGraphicsExtensions{.pdf,.jpeg,.jpg,.png}
\usepackage{adjustbox}
\usepackage{url}
\usepackage{hhline}

\aclfinalcopy % Uncomment this line for the final submission
\def\aclpaperid{88} %  Enter the acl Paper ID here

\usepackage{algorithmic}
\usepackage{algorithm}
\usepackage[tight,footnotesize]{subfigure}

\def\argmax{\operatornamewithlimits{argmax}}
\def\argmin{\operatornamewithlimits{argmin}}

\title{The Edit Distance Transducer in Action: The University of Cambridge English-German System at WMT16}

\author{Felix Stahlberg \and Eva Hasler \and Bill Byrne \\
	Department of Engineering\\
  	University of Cambridge\\
  	Cambridge, UK}

\date{}

\begin{document}

\maketitle

\begin{abstract}
This paper presents the University of Cambridge submission to WMT16. Motivated by the complementary nature of syntactical machine translation and neural machine translation (NMT), we exploit the synergies of Hiero and NMT in different combination schemes. Starting out with a simple neural lattice rescoring approach, we show that the Hiero lattices are often too narrow for NMT ensembles. Therefore, instead of a hard restriction of the NMT search space to the lattice, we propose to loosely couple NMT and Hiero by composition with a modified version of the edit distance transducer. The loose combination outperforms lattice rescoring, especially when using multiple NMT systems in an ensemble.
\end{abstract}

\section{Introduction}
\label{sec:intro}

Previous work suggests that syntactic machine translation such as Hiero~\cite{hierarchical} and Neural Machine Translation (NMT)~\cite{kalchbrenner,sutskever,choPhraseRepresentation,bahdanau} are very different and have complementary strengths and weaknesses~\cite{neubigAtwat2015,sgnmt}. Recent attempts to combine syntactic SMT and NMT report large gains over both baselines. Authors in~\cite{neubigAtwat2015} used NMT to rescore $n$-best lists which were generated with a syntax-based system. They report that even with 1000-best lists, the gains of using the NMT rescorer often do not saturate. Syntactically Guided NMT~\cite[SGNMT]{sgnmt} constrains the NMT search space to Hiero translation lattices which contain significantly more hypotheses than $n$-best lists.    In SGNMT,  an NMT beam decoder with a relatively small beam can explore spaces much larger than $n$-best lists, yielding BLEU score improvements with far fewer expensive NMT evaluations. 

However, these rescoring approaches enforce an exact match between the NMT and syntactic decoders. In general, this kind of hard restriction is best avoided when combining diverse systems~\cite{joint-decoding,pangloss}. For example, in speech recognition, ROVER~\cite{rover} is a system combination approach based on a soft voting scheme. In machine translation, minimum Bayes-risk (MBR) decoding~\cite{mbr} can be used to combine multiple systems~\cite{mbr-morph}. MBR also does not enforce exact agreement between systems as it distinguishes between the {\em hypothesis space} and the {\em evidence space}~\cite{goel2000minimum,tromble-lmbr}.

We find that Hiero lattices generated by grammars extracted with the usual heuristics~\cite{hierarchical} do not provide enough variety to explore the full potential of neural models, especially when using NMT ensembles. Therefore, we present a ``soft'' lattice-based combination scheme which uses standard operations on finite state transducers such as composition. Our method replaces the hard combination in previous methods with a similarity measure based on the edit distance, and gives the NMT decoder more freedom to diverge from the Hiero translations. We find that this loose coupling scheme is especially useful when using NMT ensembles.

\begin{figure*}[!t]
\centering
\subfigure[Standard edit distance transducer.]{\label{fig:flower}
\hspace{4em}\includegraphics[scale=0.25]{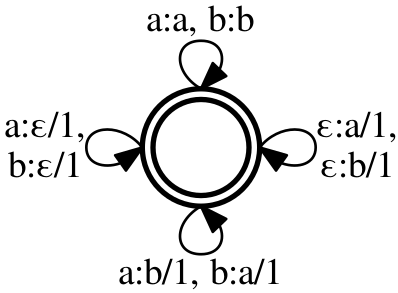}\hspace{4em}
}
\subfigure[Modified edit distance transducer $E$. `a' is an NMT OOV.]{\label{fig:modified-flower}
\hspace{4em}\includegraphics[scale=0.25]{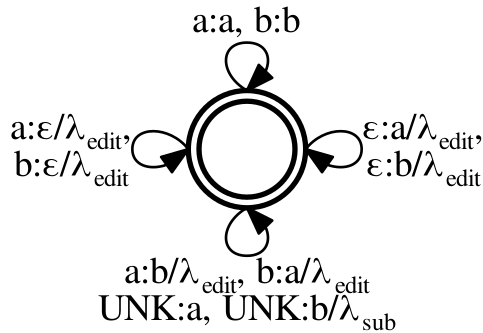}\hspace{4em}
}
\caption{``Flower automata'' for calculating edit distances over the alphabet $\{\text{a},\text{b},\text{UNK}\}$.}
\label{fig:flowers}
\end{figure*}

\section{Combining Hiero and NMT via Edit Distance Transducer}
\label{sec:combi-edit-distance}

In contrast to the strict coupling in SGNMT, we propose to loosely couple Hiero and NMT via an edit distance transducer and shortest distance search. With loose coupling, the NMT decoder is not restricted to the Hiero lattice as in previous work, but runs independently to produce translation lattices on its own, which are then combined with the Hiero lattices. The combination does not require an exact match. Instead, we will describe a procedure for combining NMT and Hiero that captures similarity under the edit distance and both the NMT and Hiero translation system scores. This scheme is implemented efficiently using standard FST operations~\cite{openfst}. First, we introduce the FST composition operation and the edit distance transducer. We will describe the whole pipeline in Sec.~\ref{ssec:loose-coupling}.

\subsection{Composition of Finite State Transducers}
\label{ssec:fstcompose}

The composition of two weighted transducers $T_1$, $T_2$ (denoted as $T_1 \circ T_2$) over a semiring $(\mathbb{K},\oplus,\otimes)$ is defined following~\cite{fstalg}

\begin{equation}
[T_1 \circ T_2](x,y)=\bigoplus_z T_1(x,z)\otimes T_2(z,y).
\end{equation}
We will make extensive use of this operation as tool for building complex automata which make use of both the NMT and Hiero translation lattices.

\subsection{The Edit Distance Transducer}
\label{ssec:edit-distance}

Composition can be used together with a ``flower automaton'' to calculate the edit distance between two sequences~\cite{mohri2003edit}. The edit distance transducer shown in Fig.~\ref{fig:flower} transduces a sequence $x$ to another sequence $y$ over the alphabet $\{\text{a},\text{b}\}$ and accumulates the number of edit operations via the transitions with cost 1. In our case, $x$ corresponds to an NMT hypothesis which is to be combined with a Hiero hypothesis $y$. In contrast to SGNMT, where we require an exact match between NMT and Hiero (up to UNKs), our edit-distance-based scheme allows different hypotheses to be combined. We replaced the standard definition of the edit distance transducer~\cite{mohri2003edit} by a finer-grained model designed to work well for combining NMT and Hiero. Instead of uniform costs, we lower the cost for UNK substitutions as we want to encourage substituting NMT UNKs by words in the Hiero translation. We distinguish between three types of edit operations.
\begin{itemize}
\item {\bf Type I}: Substituting UNK with a word outside the NMT vocabulary is free.
\item {\bf Type II}: For substitutions of UNK with a word inside the NMT vocabulary we add the cost $\lambda_{sub}$.
\item {\bf Type III}: All other edit operations are penalized with cost $\lambda_{edit}$ (and $\lambda_{edit}>\lambda_{sub}$).
\end{itemize}
We will refer to the modified edit distance transducer as $E$. Fig.~\ref{fig:modified-flower} shows $E$ over the alphabet $\{\text{a},\text{b},\text{UNK}\}$, with `a' being an NMT OOV.

\begin{figure*}[!t]
\centering
\subfigure[Example Hiero lattice $H$.]{\label{fig:hifst}\includegraphics[width=1\linewidth]{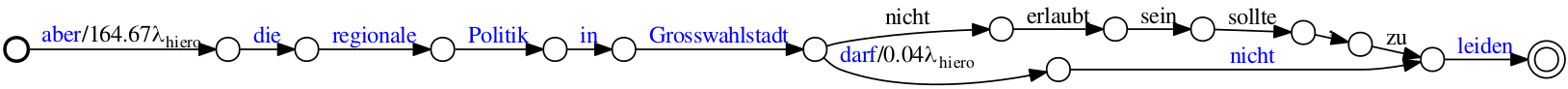}}
\subfigure[Example NMT lattice $N$.]{\label{fig:nmt}\includegraphics[width=0.8\linewidth]{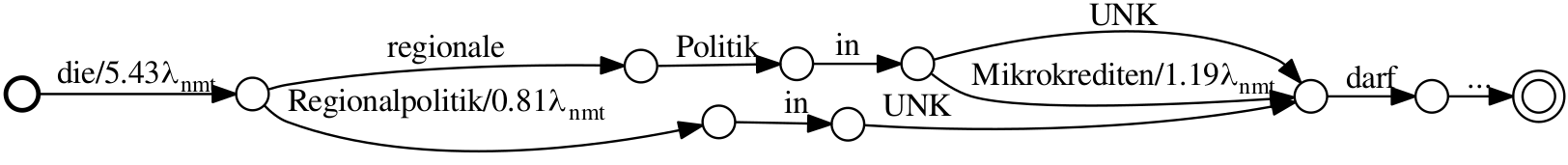}}
\subfigure[Transducer with UNK insertion arcs: $\text{Replace}(N,\mathtt{UNK},U)$.]{\label{fig:nmt_with_unk}\includegraphics[width=0.96\linewidth]{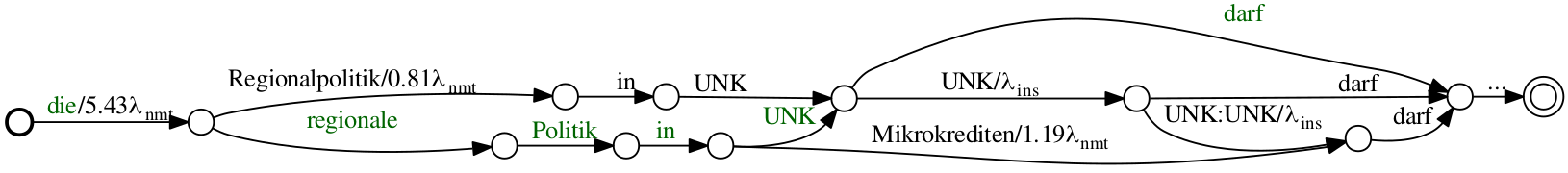}}
\subfigure[Best path in the combined transducer $C$. Hiero scores are omitted in this figure.]{\label{fig:bestpath}\includegraphics[width=1\linewidth]{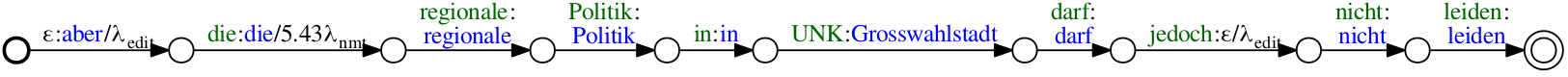}}
\subfigure[Projection of the best path: $\Pi_{UNK}(\text{ShortestPath}(C))$. The final hypothesis is {\em die regionale Politik in Grosswahlstadt darf jedoch nicht leiden}.]{\label{fig:projected}\includegraphics[width=0.93\linewidth]{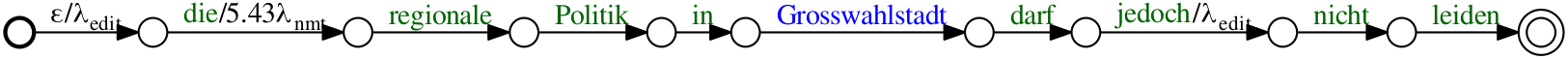}}

\caption{Combining Hiero and NMT via edit distance transducer.}
\label{fig:edit_distance_coupling}
\end{figure*}

\begin{figure}[!b]
\centering
\includegraphics[width=0.8\linewidth]{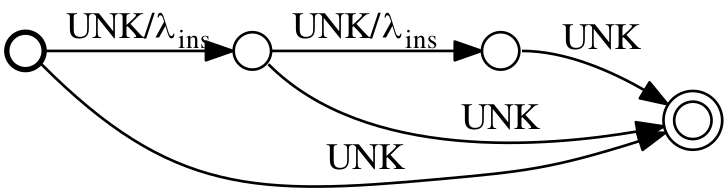}
\caption{UNK extension transducer $U$.}
\label{fig:unk_fst}
\end{figure}

\subsection{Loose Coupling of Hiero and NMT}
\label{ssec:loose-coupling}

Our edit-distance-based scheme combines an NMT translation lattice $N$ with a Hiero translation lattice $H$. Weights in $N$ and $H$ are scaled by $\lambda_{nmt}$ and $\lambda_{hiero}$, respectively. The similarity measure between NMT and Hiero translations is parametrized with $\lambda_{ins}$, $\lambda_{edit}$, and $\lambda_{sub}$. We keep the various costs separated by using transducers with tropical sparse tuple vector semirings~\cite{disambiguation}. Instead of single real-valued arc weights, this semiring uses vectors which can hold multiple features. The inner product of these vectors with a constant parameter vector determines the final weights on the arcs\footnote{The ucam-smt tutorial contains details to the tropical sparse tuple vector semiring: http://ucam-smt.github.io/tutorial/basictrans.html\#lmert\_veclats\_tst}. The sparse tuple vector semiring enables us to optimize the $\lambda$-parameters with LMERT~\cite{lmert} on a development set.

Examples for $H$ and $N$ are shown in Fig.~\ref{fig:hifst} and Fig.~\ref{fig:nmt}. The shortest path in $H$ containing the string {\em nicht erlaubt sein sollte zu} has grammatical and stylistic flaws but is complete, whereas there is a better path in $N$ with an UNK. Our goal is to merge these two hypotheses by using the NMT translation in $N$ with the UNK replaced by a word from the Hiero lattice~$H$.

\begin{enumerate}
\item {\bf Adding UNK insertions}.
We found that often NMT produces an isolated UNK token, even if multiple tokens are required. Therefore, we allow extending a single UNK token to a sequence of up to three UNK tokens. This is realized by replacing UNK arcs in $N$ with the transducer $U$ shown in Fig.~\ref{fig:unk_fst} using OpenFST's \texttt{Replace} operation. Fig.~\ref{fig:nmt_with_unk} shows the result of the replace operation when applied to the example lattice $N$ in Fig.~\ref{fig:nmt}. We denote this operation as follows:
\begin{equation}
\text{Replace}(N,\mathtt{UNK},U)
\end{equation}
\item {\bf Composition with the edit distance transducer}.
The next step finds the edit distances to the Hiero hypotheses as described in Sec.~\ref{ssec:edit-distance}.
\begin{equation}
C := \text{Replace}(N,\mathtt{UNK},U)\circ E \circ H
\end{equation}
\item {\bf Shortest path}. The above operation generates very large lattices, and dumping all of them is not feasible. We could use disambiguation~\cite{disambiguation,mohri2015disambiguation} on the combined transducer $C$ to find the best alignment for each unique NMT hypothesis. However, we only need the single shortest path in order to generate the combined translation.
\begin{equation}
\text{ShortestPath}(C)
\end{equation}
\item {\bf Projection}.
A complete path in the transducer $C$ has an NMT hypothesis on the input labels (marked green in Fig.~\ref{fig:bestpath}) and a Hiero hypothesis on the output labels (marked blue in Fig.~\ref{fig:bestpath}). Therefore, we can generate different translations from the best path in~$C$. If we project the input labels on the output labels with OpenFST's \texttt{Project}, we obtain a hypothesis $\widehat{t}_{NMT}$ in the NMT lattice $N$.

\begin{equation}
\widehat{t}_{NMT} = \Pi_1(\text{ShortestPath}(C))
\label{eq:loose-nmt}
\end{equation}

However, $\widehat{t}_{NMT}$ still contains UNKs. If we project on the input labels, we end up with the aligned Hiero hypothesis without UNKs (blue labels in Fig.~\ref{fig:bestpath})

\begin{equation}
\widehat{t}_{Hiero} = \Pi_2(\text{ShortestPath}(C))
\label{eq:loose-hiero}
\end{equation}
but we do not use the NMT translation directly. Therefore, we introduce a new projection function $\Pi_{UNK}$ which switches between preserving symbols on the input and output tapes: if the input label on an arc is UNK, we write the output label over the input label. Otherwise, we write the input label over the output label. This is equivalent to projecting the output labels to the input labels only if the input label is UNK, and then projecting the input labels to the output labels. As shown in Fig.~\ref{fig:projected}, we obtain the NMT hypothesis, but the UNK is replaced by the matching word {\em Grosswahlstadt} from the Hiero lattice. Thus, the final combined translation is described by the following term:
\begin{equation}
\widehat{t}_{comb} = \Pi_{UNK}(\text{ShortestPath}(C))
\label{eq:loose}
\end{equation}
\end{enumerate}

In general, the final hypothesis $\widehat{t}_{comb}$ is a mix of an NMT and a Hiero hypothesis. We do not search for $\widehat{t}_{comb}$ directly but for pairs of NMT and Hiero translations which optimize the individual model scores as well as the distance between them. Stated more formally, the shortest path in $C$ yields a pair $(\widehat{t}_{NMT},\widehat{t}_{Hiero})$ for which holds
\begin{equation}
\label{eq:non-fst}
\begin{aligned}
\widehat{t}_{NMT},\widehat{t}_{Hiero} =\argmin_{(t_{N},t_{H})\in N \times H} \Big( d_{edit}(t_{N},t_{H})\\
 +\lambda_{nmt}\cdot S_{N}(t_{N}|s)+\lambda_{hiero}\cdot S_{H}(t_{H}|s)\Big)
\end{aligned}
\end{equation}
where $d_{edit}(t_{N},t_{H})$ is the modified edit distance between $t_{N}$ and $t_{H}$ (according $E$ and $U$), and $S_{N}(t_{N}|s)$ and $S_{H}(t_{H}|s)$ are the scores NMT and Hiero assign to the translations given source sentence $s$. If we interpret these scores as negative log-likelihoods, we arrive at a probabilistic interpretation of Eq.~\ref{eq:non-fst}.

\begin{equation}
\label{eq:non-fst-prob}
\begin{aligned}
\widehat{t}_{NMT},\widehat{t}_{Hiero} &=\argmax_{(t_{N},t_{H})\in N \times H} \Big(\\
& \mathrm{e}^{-d_{edit}(t_{N},t_{H})}\cdot P(t_{N},t_{H}|s) \Big) \\
\end{aligned}
\end{equation}
with (assuming independence)
\begin{equation*}
P(t_{N},t_{H}|s):=P_{N}(t_{N}|s)^{\lambda_{nmt}}\cdot P_{H}(t_{H}|s)^{\lambda_{hiero}}.
\end{equation*}

Eq.~\ref{eq:non-fst-prob} suggests that we maximize the product of two quantities -- the similarity between Hiero and NMT hypotheses and their joint probability. The FST operations allow to optimize over the set $N \times H$ efficiently. Note that the NMT lattice $N$ is rather small in our case ($|N|\leq 20$) due to the small beam size used in NMT decoding. This makes it possible to solve Eq.~\ref{eq:non-fst} almost always without pruning~\footnote{We limit the Hiero lattices to a maximum of 100,000 nodes with OpenFST's \texttt{Prune} to remove the worst outliers.}.

\section{Experimental Setup}
\label{sec:setup}

The parallel training data includes {\em Europarl v7},  {\em Common Crawl}, and  {\em News Commentary v10}. 
Sentence pairs with sentences longer than 80 words or length ratios exceeding  2.4:1 were deleted, as were {\em Common Crawl} sentences from other languages~\cite{langDetect}. 
We use {\em news-test2014} (the filtered version) as a development set, and keep {\em news-test2015} and {\em news-test2016} as test sets.

The NMT systems are built using the  Blocks framework~\cite{blocks} based on the Theano library~\cite{theano} with the network architecture and hyper-parameters as in~\cite{bahdanau}: the encoder and decoder networks consist of 1000~gated recurrent units~\cite{choPhraseRepresentation}. The decoder uses a single maxout~\cite{maxout} output layer with the feed-forward attention model described in~\cite{bahdanau}. In our final ensemble, we use 8 independently trained NMT systems with vocabulary sizes between 30,000 and 60,000.

Rules for our En-De Hiero system were extracted as described in~\cite{hifst-grammar}. A 5-gram language model for the Hiero system was trained on  WMT16 parallel and monolingual data~\cite{Heafield-estimate}. 

We apply gentle post-processing to the German output for fixing small number and currency formatting issues. The English source sentences in the training corpus are lower-cased. During decoding, we lower case only in-vocabulary words, and pass through OOVs with correct casing. We apply a simple heuristic for recognizing surnames to avoid literal translation of them into German\footnote{We mark a word as surname if it has occurred after a first name, is on a census list of known surnames, and is written with a capitalized initial letter.}.

\begin{table*}
\small
\centering

\begin{tabular}{|ll|r|r|r|}
\hline
 \textbf{Setup} && \textbf{news-test2014} & \textbf{news-test2015} & \textbf{news-test2016} \\ \hline
 \multicolumn{2}{|l|}{Best in competition\footnotemark}  & 20.6 & 25.2 & 34.8 \\  \hhline{|==|=|=|=|}
 \multicolumn{2}{|l|}{Hiero baseline}  & 18.9 & 21.2 & 26.0 \\  \hline
Single NMT & Pure NMT & 17.5 & 19.6 & 23.2 \\
 & SGNMT (lattice rescoring) & 21.2 & 23.5 & 28.7 \\
 & Edit distance transducer based combination & 21.7 & 24.1 & 28.6 \\ \hline
Ensemble NMT & Pure NMT & 19.4 & 21.7 & 25.4 \\
 & SGNMT (lattice rescoring) & 21.9 & 24.6 & 29.7 \\
 & Edit distance transducer based combination & {\bf 22.9} & {\bf 25.7} & {\bf 31.3} \\
\hline
\end{tabular}
\caption{\label{tab:results} English-German lower-cased BLEU scores calculated with Moses \texttt{mteval-v13a.pl}.}
\end{table*}

\begin{table}[t!]
\small
\centering
\begin{tabular}{|l|r|}
\hline
{\bf Method} & {\bf BLEU} \\
\hline
NMT baseline: $\text{ShortestPath}(N)$ & 25.4 \\
Hiero baseline: $\text{ShortestPath}(H)$ & 26.4 \\
\hline
NMT hypothesis used for combination: $\widehat{t}_{NMT}$ & 26.7 \\
Hiero hypothesis used for combination: $\widehat{t}_{Hiero}$ & 30.4 \\
Combined translation: $\widehat{t}_{comb}$ & 31.3 \\
\hline
\end{tabular}
\caption{\label{tab:projection} Projection methods on {\em news-test2016} with NMT 8-ensemble. }
\end{table}

\section{Results}
\label{sec:results}

\begin{table*}
\small
\centering
\begin{tabular}{|l|r|r|}
\hline
{\bf Distance measure component} & {\bf Avg.\ number per sentence} & {\bf Percentage of affected sentences} \\
\hline
UNK insertions ($U$) & 0.16 & 12.9\% \\
UNK$\rightarrow$non-OOV substitutions (Type II) & 1.34 & 55.9\% \\
Other edit operations (Type III) & 1.74 & 61.7\% \\
\hline
\end{tabular}
\caption{\label{tab:alignment} Breakdown of the distances measured between NMT and Hiero along the shortest path in $C$ on {\em news-test2016}.}
\end{table*}
% in lattices: nmt,edit,ins-unk,unk-to-in-vocab,hifst

\footnotetext{http://matrix.statmt.org/}

Tab.~\ref{tab:results} reports performance on {\em news-test2014}, {\em news-test2015}, and {\em news-test2016}\footnote{The code we used for SGNMT and ensembling is available at http://ucam-smt.github.io/sgnmt/html/.}. Similarly to previous work~\cite{sgnmt}, we observe that rescoring Hiero lattices with NMT (SGNMT) outperforms both NMT and Hiero baselines significantly on all test sets. For SGNMT, we see further improvements of between +0.7 BLEU ({\em news-test2014}) and +1.1 BLEU ({\em news-test2015}) by using NMT ensembles rather than single NMT. However, these gains are rather small considering the improvements from using ensembles for the (pure) NMT baseline (between +1.9 BLEU and +2.2 BLEU). Our combination scheme makes better use of the ensembles. We report 31.3 BLEU on {\em news-test2016}, which  in the English-German WMT'16 evaluation is among the best systems (within 0.1 BLEU) which do not use back-translation~\cite{backtranslation}. Back-translation is a technique for making use of monolingual data in NMT training, and we expect our system could benefit from back-translation, although we leave this analysis to future work.

The combination procedure we propose is non-trivial.  It is not immediately clear how the gains arise as the final scores are mixtures between edit distance costs, NMT scores, and Hiero scores.  In the remainder we will try to provide some insight. Unless stated otherwise, we report investigations into the Hiero + NMT 8-system ensemble which yields the best results in Tab.~\ref{tab:results}.   

\begin{figure}[b!]
\centering
\includegraphics[width=1\linewidth]{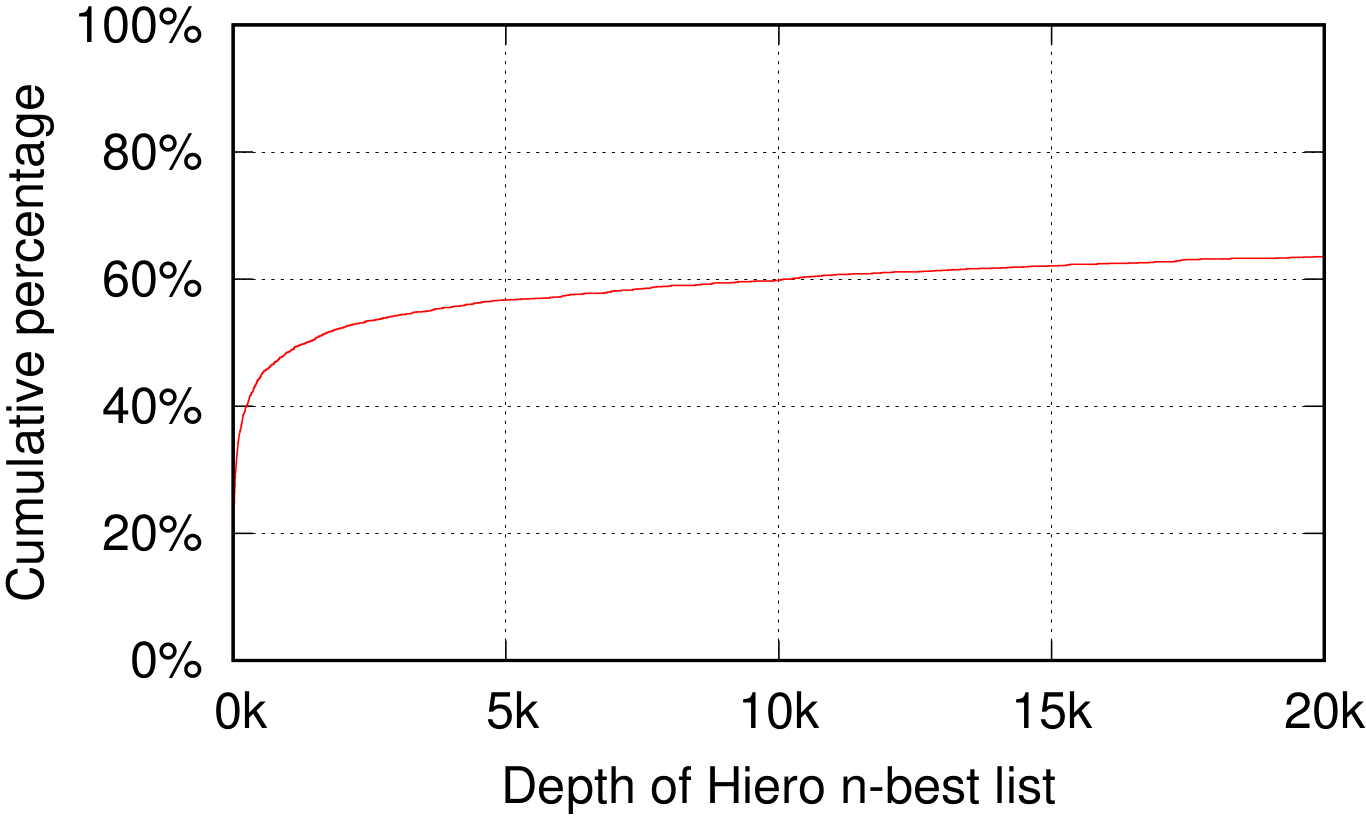}
\caption{Percentage of $\widehat{t}_{Hiero}$ hypotheses found in the baseline Hiero $n$-best list.}
\label{fig:nbest-lookup}
\end{figure}

First, we focus on the projection function $\Pi_{UNK}(\cdot)$ which switches between preserving the input and output label at the UNK symbol to produce the combined translation $\widehat{t}_{comb}$ (Eq.~\ref{eq:loose}). As explained in Sec.~\ref{ssec:loose-coupling}, we can use OpenFST's \texttt{Project} operation to fetch the NMT and Hiero hypotheses $\widehat{t}_{NMT}$ and $\widehat{t}_{Hiero}$ which have been used to produce the combined translation (Eq.~\ref{eq:loose-nmt} and \ref{eq:loose-hiero}). Tab.~\ref{tab:projection} shows that the hypotheses that are aligned in the final transducer are often not the 1-best translations of any of the baseline systems. Remarkably, using the $\widehat{t}_{Hiero}$ translations results in 30.4 BLEU, which is a very substantial improvement over the baseline Hiero system (26.0 BLEU). Note that this BLEU score is achieved with hypotheses from the original Hiero lattice $H$ but weighted in combination with the NMT scores and the edit distance. However, these selected paths are often given very low scores by Hiero: in only 8.6\% of the sentences is the Hiero hypothesis left unchanged. If we look for $\widehat{t}_{Hiero}$ in the Hiero $n$-best list, we find that even very deep 20,000-best lists contain only 63.5\% of the Hiero hypotheses which were selected by the combination scheme (Fig.~\ref{fig:nbest-lookup}). This indicates the benefit in using lattice-based approaches over $n$-best lists.

Next, we investigate the distance measure between NMT and Hiero translations, which is realized with the UNK insertion transducer $U$ and the modified edit distance transducer $E$ (Sec.~\ref{ssec:loose-coupling}). Tab.~\ref{tab:alignment} shows that UNK insertions are relatively rare compared to the edit operations of types II and III allowed by $E$ (Sec.~\ref{ssec:loose-coupling}). The average edit distance between NMT and Hiero disregarding UNKs on the best path (type III) is 1.74. In 61.7\% of the cases the input and output labels differ not only at UNK -- i.e.\ in only 38.3\% of the sentences do we have an exact match between NMT and Hiero. We note that UNK is often replaced with an NMT in-vocabulary word (55.9\% of the sentences). It seems that NMT often produces an UNK even if a better word is in the NMT vocabulary. This could be due to the over-representation of UNK in the NMT training corpus.

To study the effectiveness of our edit distance transducer based combination scheme in correcting NMT UNKs, we trained individual NMT systems with vocabulary sizes between 10,000 and 60,000. Tab.~\ref{tab:voc-size} shows that nearly one in six tokens (16.3\%) produced by our pure NMT system with a vocabulary size of 30,000 are UNKs. Increasing the NMT vocabulary to 50k or 60k does improve pure NMT very significantly, but results show that these improvements are already captured by the combination scheme with Hiero. As in the literature, we see large variation in performance over individual NMT systems even with the same vocabulary size~\cite{bpe}, which could explain the small performance drop when increasing the vocabulary size from 50k to 60k.

\begin{table}[t!]
\small
\centering
\begin{tabular}{|r|r|r|r|}
\hline
{\bf Vocabulary} & \multicolumn{2}{c|}{{\bf Pure NMT}} & {\bf NMT+Hiero} \\
\multicolumn{1}{|c|}{{\bf size}} & {\bf BLEU} & {\bf \# of UNKs} & \multicolumn{1}{c|}{{\bf BLEU}} \\
\hline
10,000 & 18.9 & 18.0\%  & 28.1 \\
30,000 & 21.6 & 16.3\%   & 28.8 \\
50,000 & 23.2 & 9.1\%   & 28.6 \\
60,000 & 22.9 & 9.9\%  & 28.5 \\
\hline
\end{tabular}
\caption{\label{tab:voc-size} BLEU scores on {\em news-test2016} for different vocabulary sizes (single NMT). Each individual NMT system is combined with Hiero as described in Sec.~\ref{ssec:loose-coupling}.}
\end{table}

\begin{figure}[t!]
\centering
\includegraphics[width=1\linewidth]{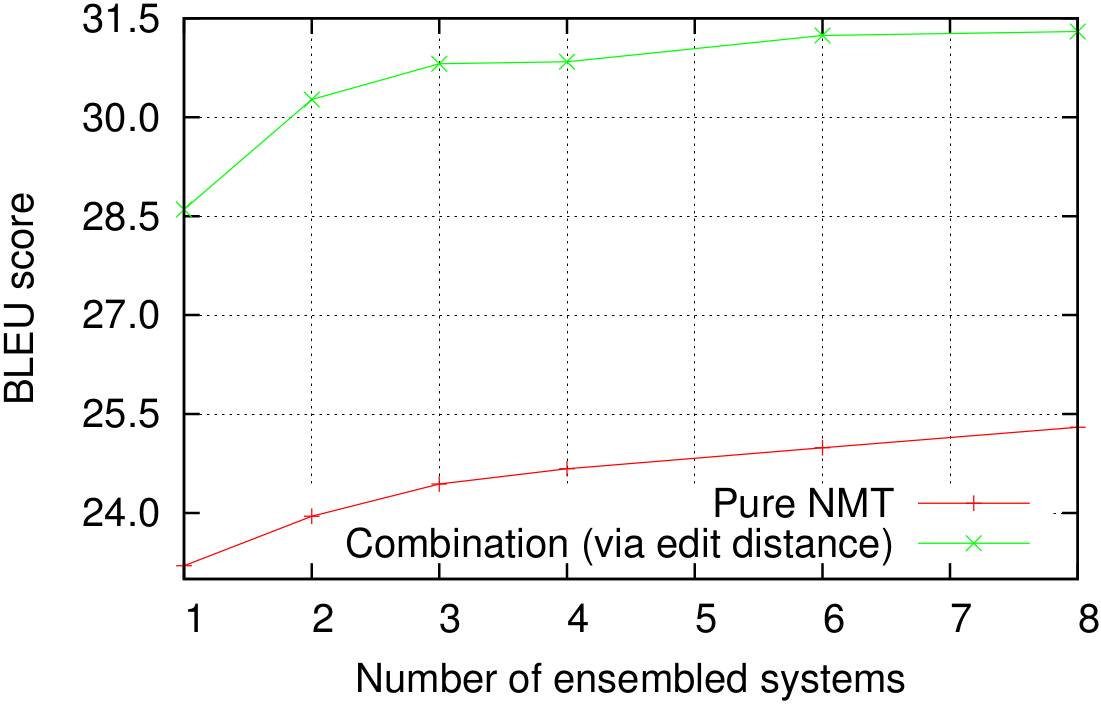}
\caption{BLEU score over the number of systems in the ensemble on {\em news-test2016}.}
\label{fig:ensemble-size}
\end{figure}

One important practical issue for system building is the number of systems to be ensembled as training each individual NMT system takes a significant amount of time. Fig.~\ref{fig:ensemble-size} indicates that even for 8-ensembles the gains for pure NMT do not seem to saturate. The combination with Hiero via edit distance transducer also greatly benefits from using ensembles, but most of the gains are gotten with fewer systems.

\section{Conclusion and Future Work}
\label{sec:conclusion}

We have presented a method based on the edit distance that is effective in combining Hiero SMT systems with NMT ensembles.  Our approach makes use of standard WFST operations, and we showed the effectiveness of the approach with a successful WMT'16 submission for English-German.
In the future, we are planning to add back-translation~\cite{backtranslation} and investigate the use of character- or subword-based NMT~\cite{bpe,chitnis-denero:2015:EMNLP,char-nmt,no-seg-nmt,hybrid-nmt} within our combination framework.

\section*{Acknowledgements}
This work  was  supported  by the U.K. Engineering and Physical Sciences Research Council (EPSRC grant EP/L027623/1).

\bibliography{refs}
\bibliographystyle{acl2016}

\end{document}